\documentclass[10pt,twocolumn,letterpaper]{article}

\usepackage[pagenumbers]{cvpr}

\usepackage{nicefrac}
\usepackage{microtype}
\usepackage{wrapfig}
\usepackage{acro}
\usepackage{makecell}
\usepackage{multirow}
\usepackage{float}

\DeclareAcronym{kgfp}{
  short = KGFP,
  long  = Knowledge-Guided Failure Prediction
}

\DeclareAcronym{ood}{
  short = OOD,
  long  = Out-of-Distribution
}
\DeclareAcronym{id}{
  short = ID,
  long  = In-Distribution
}

\DeclareAcronym{fpr}{
  short = FPR,
  long  = False Positive Rate
}
\DeclareAcronym{tpr}{
  short = TPR,
  long  = True Positive Rate
}
\DeclareAcronym{auroc}{
  short = AUROC,
  long  = Area Under ROC Curve
}
\DeclareAcronym{roc}{
  short = ROC,
  long  = Receiver Operating Characteristic
}
\DeclareAcronym{auc}{
  short = AUC,
  long  = Area Under Curve
}
\DeclareAcronym{iou}{
  short = IoU,
  long  = Intersection over Union
}

\DeclareAcronym{msp}{
  short = MSP,
  long  = Maximum Softmax Probability
}
\DeclareAcronym{knn}{
  short = KNN,
  long  = K-Nearest Neighbors
}
\DeclareAcronym{vim}{
  short = ViM,
  long  = Virtual-logit Matching
}
\DeclareAcronym{pca}{
  short = PCA,
  long  = Principal Component Analysis
}

\DeclareAcronym{mlp}{
  short = MLP,
  long  = Multi-Layer Perceptron
}
\DeclareAcronym{bce}{
  short = BCE,
  long  = Binary Cross-Entropy
}
\DeclareAcronym{fpn}{
  short = FPN,
  long  = Feature Pyramid Network
}
\DeclareAcronym{vit}{
  short = ViT,
  long  = Vision Transformer
}
\DeclareAcronym{gelu}{
  short = GELU,
  long  = Gaussian Error Linear Unit
}

\DeclareAcronym{lars}{
  short = LARS,
  long  = Layer-wise Adaptive Rate Scaling
}
\DeclareAcronym{hdf}{
  short = HDF5,
  long  = Hierarchical Data Format 5
}

\definecolor{cvprblue}{rgb}{0.21,0.49,0.74}
\usepackage[pagebackref,breaklinks,colorlinks,allcolors=cvprblue]{hyperref}

\def\confName{CVPR}
\def\confYear{2026}

\title{Knowledge-Guided Failure Prediction: \\ Detecting When Object Detectors Miss Safety-Critical Objects\thanks{Accepted at the SAIAD Workshop, CVPR 2026.}}

\author{Jakob Paul Zimmermann\\
Fraunhofer HHI\\
Berlin, Germany\\
{\tt\small jakob.zimmermann@campus.tu-berlin.de}
\and
Gerrit Holzbach\\
Fraunhofer IOSB\\
Karlsruhe, Germany\\
{\tt\small gerrit.holzbach@iosb.fraunhofer.de}
\and
David Lerch\\
Fraunhofer IOSB\\
Karlsruhe, Germany\\
{\tt\small david.lerch@iosb.fraunhofer.de}
}

\begin{document}
\maketitle

\begin{abstract}
Object detectors deployed in safety-critical environments can fail silently, e.g. missing pedestrians, workers, or other safety-critical objects without emitting any warning. Traditional \ac{ood} detection methods focus on identifying unfamiliar inputs, but do not directly predict functional failures of the detector itself. We introduce \ac{kgfp}, a representation-based monitoring framework that treats missed safety-critical detections as anomalies to be detected at runtime. \Ac{kgfp} measures semantic misalignment between internal object detector features and visual foundation model embeddings using a dual-encoder architecture with an angular distance metric. A key property is that when either the detector is operating outside its competence or the visual foundation model itself encounters novel inputs, the two embeddings diverge, producing a high-angle signal that reliably flags unsafe images. We compare our novel \ac{kgfp} method to baseline \ac{ood} detection methods. On COCO person detection, applying \ac{kgfp} as a selective-prediction gate raises person recall among \emph{accepted} images from 64.3\% to 84.5\% at 5\% \ac{fpr}, and maintains strong performance across six COCO-O visual domains, outperforming \ac{ood} baselines by large margins. Our code, models, and features are published at \url{https://gitlab.cc-asp.fraunhofer.de/iosb_public/KGFP}
\end{abstract}

\acresetall

\section{Introduction}

Object detectors power safety-critical applications from autonomous driving to video surveillance, yet they fail unpredictably when encountering novel or challenging visual conditions~\citep{hoiem2012diagnosing, hendrycks2021natural}.
As automation increases, such perception components are deployed more widely in operational decision loops and therefore increasingly fall under safety certification expectations in the sense of systematic AI systems engineering~\citep{pfrommer2022ki}.

A pedestrian detection system trained on clear weather may miss occluded pedestrians in fog; a construction site monitor~\citep{Hagmanns2025a} may fail to detect workers in unusual poses.

\begin{figure}[H]
\centering
\includegraphics[width=\columnwidth]{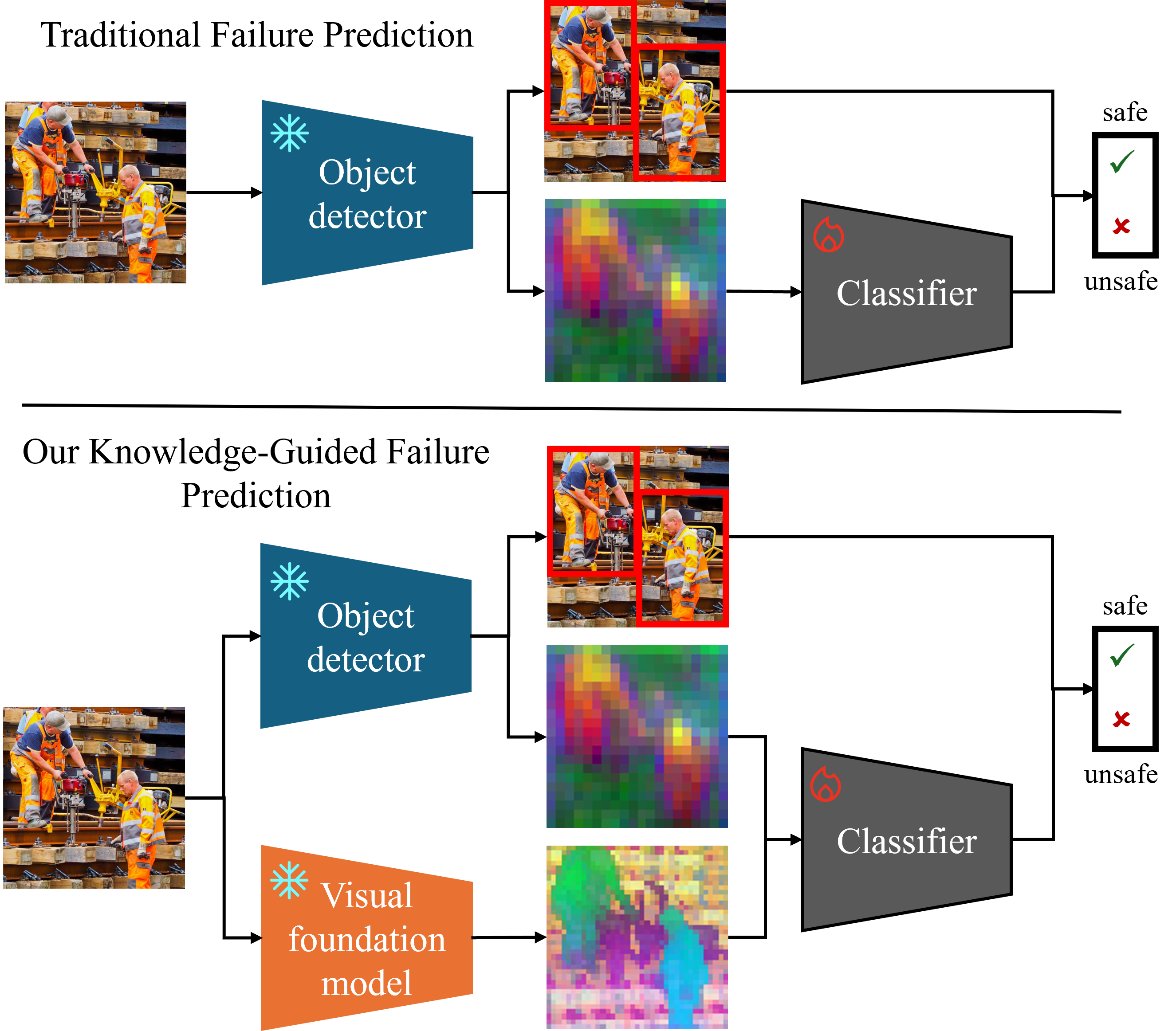}
\caption{Overview of our proposed knowledge-guided failure prediction for object detectors. Traditional failure prediction methods including OOD utilize only internal features of the object detector. We leverage visual foundation model features to classify if an image is safe for person detection.}
\label{fig:concept}
\end{figure}

These \textbf{silent failures}—where detectors confidently return empty results despite objects present—can have catastrophic consequences.

Traditional approaches to detector robustness focus on \textit{improving detection accuracy} through domain adaptation~\citep{zhang2018occluded}, data augmentation, or architecture refinement~\citep{lin2017feature}. However, even state-of-the-art detectors exhibit failure modes including missing objects in complex contexts~\citep{sun2017seeing} and under novel visual conditions~\citep{yang2021introspective}.
For high-assurance settings, achieving the evidence and risk reduction expected at Safety Integrity Level 4 ~\citep{IEC61508:2010} cannot rely on test set accuracy alone; it also requires mechanisms that detect and control unsafe behavior at runtime when assumptions are violated.
Rather than attempting to eliminate failures entirely, \textbf{runtime monitoring}~\citep{schotschneider2025runtime, kang2018model} aims to detect anomalous situations in which the perception system is likely to miss safety-critical objects and to issue timely alerts.
Such alerts are essential for the system to \textbf{adapt and operate} safely under uncertainty, enabling downstream controllers to trigger fallback policies (e.g., initiating a minimal risk maneuver or querying a human operator) when functional failures are predicted.

We formulate \textbf{failure prediction for safety-critical objects} as a supervised binary classification task: given an image and a detector's predictions, determine whether the detector has missed any safety-critical objects. Crucially, we focus on a \textbf{safety-critical subset of object classes}—for pedestrian detection systems, persons are the primary safety concern; objects like small birds or distant vehicles matter less for immediate safety. This differs fundamentally from the typically unsupervised binary decision problem of \ac{ood} detection~\citep{cui2022generalized}.

\Ac{kgfp} is trained to predict the safety of \ac{id} data, and demonstrates generalization capabilities to \ac{ood} data.

A key challenge in anomaly-based monitoring is alarm fatigue. Unlike methods that flag any novel input, \ac{kgfp} focuses on functional failures, significantly reducing false alarms by ignoring benign novelty that does not impact the detection of safety-critical objects.
Moreover, unlike conventional \ac{ood} detection, it flags \ac{id} images where the detector is struggling.

Recent foundation models trained on billions of images~\citep{radford2021clip} capture rich semantic knowledge beyond task-specific detectors. Self-supervised vision transformers like DINO~\citep{caron2021dino} learn particularly powerful semantic representations. Our key insight: \textbf{semantic similarity between learned encodings of detector internal features and foundation model embeddings indicates detection reliability}. When encoded representations of YOLOv8's internal activations align with encoded DINO semantic embeddings, the detector operates within its competence zone. Misalignment signals potential failure.

Our main contributions are as follows:
\begin{enumerate}
    \item \textbf{Safety-focused failure prediction}: To the best of our knowledge, we introduce the first framework that explicitly predicts detector failures with respect to safety-critical object classes (persons) rather than generic distribution membership, enabling targeted monitoring of safety-critical detections. We propose a novel evaluation metric that quantifies the percentage of ground-truth persons detected in accepted images (Person Recall) at a 5\% false-alarm rate (\ac{fpr}) of the \ac{kgfp} module.
    \item \textbf{Foundation model integration}: We demonstrate that self-supervised world-knowledge (DINO) improves failure prediction over detector features alone through multi-scale fusion with cross-attention. We propose a dual-encoder architecture that uses cosine similarity between encoded representations of YOLOv8 internal features and DINO embeddings, where angular divergence directly signals unsafe images in which safety-critical objects will be missed.
    \item \textbf{Comprehensive evaluation}: We perform systematic ablations across architecture, training, and foundation model choices with strong baselines on both \ac{id} data and novel visual domains.
\end{enumerate}

\section{Related Work}

\subsection{\Acl{ood} Detection}

\Ac{ood} detection aims to identify inputs from unknown distributions~\citep{cui2022generalized, yang2021generalized}. Classification-based methods leverage model outputs: \ac{msp}~\citep{hendrycks2017baseline}, ODIN~\citep{liang2018enhancing} with temperature scaling, Energy scores~\citep{liu2020energy}, and activation-based methods like ReAct~\citep{sun2021react, djurisic2023activation}. Distance-based approaches measure feature space deviations: Mahalanobis distance~\citep{lee2018simple}, \ac{knn}~\citep{sun2022knn}, and recent improvements via feature normalization~\citep{muller2025mahalanobis, park2023feature}.
GRAM~\citep{sastry2020gram} computes Gram matrices (channel covariance) from YOLOv8 feature maps at each scale using orders 1--5, fits min/max statistics per scale on training data, and flags test images as unsafe when Gram matrix elements fall outside these bounds. We use no spatial pooling to preserve full feature map statistics. \Ac{knn}~\citep{sun2022knn} stores spatially-pooled YOLOv8 features from training images. At test time, it computes the Euclidean distance to the 5th nearest training neighbor per scale; high distances indicate anomalous inputs and potential failure. We use L2 normalization for stable distance computation.
\Ac{vim}~\citep{wang2022vim} fits \ac{pca} with 100 components per YOLOv8 scale on training features and projects test features onto the residual subspace (orthogonal to the principal components). Large residual norms indicate \ac{ood} samples, i.e.\ features that deviate from the low-dimensional \ac{id} manifold.
We also trained a DINO-ViM variant, where we use DINO embeddings instead of internal YOLOv8 activations.

Although all of these methods excel at image classification \ac{ood} detection, \textbf{object detection presents unique challenges}: (1) spatial localization requirements, (2) multiple objects per image, (3) partial detection scenarios. Recent work adapts object detectors for OOD tasks \citep{zolfi2024yolood}, but focuses on flagging novel concepts in the scene rather than predicting safety-critical failures (e.g., missed pedestrians).

Angular margins have proven effective for learning discriminative embeddings in face recognition~\citep{deng2019arcface, meng2021magface, boutros2022elastic}.
Recent work extends angular margins to \ac{ood} detection~\citep{tatachar2025angular}, showing that cosine-based similarity naturally separates \ac{id} from \ac{ood} samples when features are normalized. Our approach builds on these insights, using cosine similarity between dual-encoder projections as a safety metric, where angular distance directly measures semantic alignment between detector and foundation model representations.

\subsection{Runtime Monitoring for Safety}

Runtime monitoring provides safety assurance for deployed ML systems~\citep{schotschneider2025runtime}. Comprehensive frameworks have been developed for perception systems: ~\citet{ferreira2024safety} survey threats (\ac{id} errors, novel visual conditions, adversarial attacks) and detection mechanisms, while ~\citet{tran2025safety} demonstrate simulation-based verification using Linear Temporal Logic for autonomous driving. Model assertions~\citep{kang2018model} check intermediate activations against expected distributions. Recent work by ~\citet{torpmannhagen2025runtime} proposes a paradigm shift from binary \ac{ood} classification to regression-based loss prediction, training Generalized Additive Models to directly predict task loss (cross-entropy, Jaccard index) from \ac{ood}-ness features (\ac{msp}, Energy, \ac{knn} distance) for continuous risk assessment. While loss prediction provides continuous risk quantification, it relies on proxy regression rather than directly predicting safety-critical events (missed persons). Our binary formulation enables end-to-end training with explicit safety labels, learning semantic features that directly indicate when safety-critical objects are missed.

For object detection specifically, \citet{yang2021introspective} predict false negatives using detector uncertainty, while ~\citet{yatbaz2024runtime} monitor early layer patterns in 3D detectors. ~\citet{he2025mitigating} address YOLOv8 hallucinations on \ac{ood} inputs through proximal \ac{ood} fine-tuning. Most closely related, DECIDER~\citep{subramanyam2024decider} trains a separate classifier to predict whether a detector will fail on a given input, but does not incorporate foundation model knowledge or angular distance metrics. However, these approaches rely solely on detector-internal signals. Our work differs by incorporating external semantic knowledge from foundation models, enabling detection of failures where internal features appear normal but semantic misalignment indicates unreliable predictions.

Unlike DECIDER \cite{Subramanyam2024_decider}, which requires computing similarity against class-specific textual attribute embeddings during inference to generate an auxiliary model's predictions, our framework relies solely on the angular alignment with general-purpose visual foundation model features, eliminating the dependency on predefined text definitions.

\subsection{Foundation Models for Robustness}

Self-supervised image and video models like DINO~\citep{caron2021dino} and DINOv2~\citep{oquab2024dinov2} or vision-language models like CLIP~\citep{radford2021clip} learn transferable representations from massive unlabeled data. Recent work applies these to zero-shot anomaly detection: WinCLIP~\citep{jeong2023winclip}, AnomalyCLIP~\citep{zhou2024anomalyclip}, and DINO prototypes~\citep{sinhamahapatra2025finding}. ~\citet{wang2025dinoyolo} fuse DINO with YOLOv8 for data-efficient detection. Broader surveys~\citep{cao2024visual} examine foundation models for visual anomaly detection across industrial and medical domains.
Our work differs critically: while Wang et al. enhance detection accuracy, \textbf{we leverage foundation models for failure prediction}.

Modern detectors employ multi-scale feature pyramids~\citep{lin2017feature}: YOLO~\citep{redmon2016yolo} predicts at multiple resolutions, Faster R-CNN~\citep{ren2015faster} uses Region Proposal Networks, and DETR variants~\citep{caron2021dino} apply transformers. ~\citet{hoiem2012diagnosing} analyze detector failures systematically, finding size and occlusion as dominant factors.

\section{Method}
We present \ac{kgfp}, a supervised monitoring framework that predicts when an object detector will miss safety-critical objects. Unlike standard \ac{ood} detection, which flags distributional novelty in an unsupervised manner, \ac{kgfp} is trained with explicit safe/unsafe labels derived from detector performance, enabling it to distinguish harmful failures from benign distribution shifts.

\subsection{Problem Formulation}

Given an image $\mathbf{x}$ and object detector $D$, let $\mathcal{Y}_{\text{safe}} = \{y_1, \ldots, y_n\}$ denote ground-truth bounding boxes for \textbf{safety-critical object classes}. For example, in pedestrian detection systems, $\mathcal{Y}_{\text{safe}}$ contains only persons, ignoring all other object classes. The detector produces predictions $\hat{\mathcal{Y}} = D(\mathbf{x})$ across all classes. Define the \textbf{failure label}:
\begin{equation}
    z(\mathbf{x}) = \begin{cases}
        \multirow{2}{*}{1} & \text{(unsafe) if any } y_i \in \mathcal{Y}_{\text{safe}} \\ & \text{ undetected (\ac{iou}} < 0.5\text{)} \\
        \multirow{2}{*}{0} & \text{(safe) if all safety-critical} \\ 
        & \text{objects matched}
    \end{cases}
\end{equation}

The failure prediction task: predict $z(\mathbf{x})$ from $(D(\mathbf{x}), \mathbf{x})$ to identify unsafe images where $D$ will miss safety-critical objects. This formulation differs from \ac{ood} detection by directly measuring \textit{functional failure on safety-critical objects} rather than distributional novelty. 

The objective of this study is to develop a model capable of predicting the failure of the object detector on a test sample. The technical approach entails the identification of this issue as OOD detection. In this context, the ID samples are correctly classified, while the detector exhibits failure on OOD data.
Our training and evaluation is exclusively focused on the person class, as pedestrian detection serves as the primary motivation for safety-critical applications.

Conceptually, the default output of a safety monitor for random input should be a safety warning. As two high dimensional random vectors are close to orthogonal with high probability~\citep{vershynin2018high}, this expected behavior is built into the concept of \ac{kgfp}.

\subsection{Architecture: Dual-Encoder with Angular Metric}

\begin{figure*}[t]
\centering
\includegraphics[width=\textwidth]{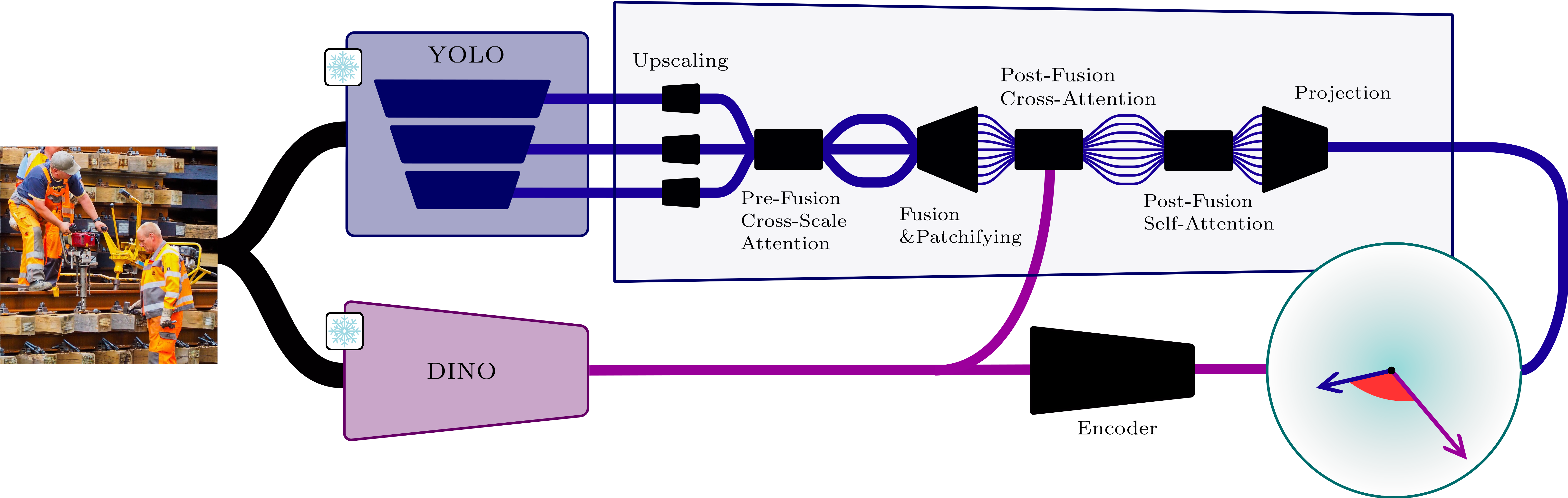}
\caption{The dual-encoder architecture of our \ac{kgfp}. We use pretrained DINO and YOLO models as backbones. For our \ac{kgfp} we freeze the pretrained backbones and fine-tune a fusion framework. During evaluation the distance between DINO features and the fused features serves as the measure for our failure prediction.}
\label{fig:architecture}
\end{figure*}

In order to evaluate a specialized model like an object detector we leverage the visual foundation model DINO~\cite{caron2021dino}.
We introduce a dual-encoder architecture that processes multi-scale YOLOv8~\cite{jocher2023yolov8} \ac{fpn} activations through cross-scale fusion and transformer blocks with self-attention and cross-attention to foundation model embeddings (see Figure~\ref{fig:architecture}). The resulting detector and world-knowledge representations are projected onto a shared embedding space, where their angular similarity provides a direct measure of detection reliability. In our ablation studies we show that our designed angular similarity metric performs on par on OOD and outperforms the MLP baseline on ID setting. This underscores the effectiveness of our angular failure metric.

\paragraph{Multi-Scale Feature Extraction}
For each image $\mathbf{x}$, we extract two complementary representations:
\begin{itemize}
    \item \textbf{Predictor features}: YOLOv8l~\citep{jocher2023yolov8} (large model) produces multi-scale internal features from its \ac{fpn} at levels $\{P3, P4, P5\}$ with channel dimensions $\{256, 512, 512\}$ and spatial resolutions $80\!\times\!80$, $40\!\times\!40$, $20\!\times\!20$ respectively (at 640$\times$640 input). These scales correspond to detection of small (P3), medium (P4), and large (P5) objects. All scales are projected to a common channel dimension and upsampled to the largest spatial resolution ($80\!\times\!80$, matching P3).
    \item \textbf{World-knowledge features} $f_{wk} \in \mathbb{R}^{768}$: DINO~\citep{caron2021dino} \ac{vit} global [CLS] embeddings extracted from the input image resized to $518\!\times\!518$~\citep{oquab2024dinov2}.
\end{itemize}

The overall architecture is shown in Figure~\ref{fig:architecture}.

\paragraph{Pre-Fusion Cross-Scale Attention}
Before fusing YOLO's multi-scale features, we apply cross-scale attention to allow information exchange between pyramid levels. Each scale attends to features from other scales, where each scale is treated as a single token. This allows the model to create scale-aware representations where, for example, P3 features (small objects) can leverage context from P5 features (scene-level patterns). The three scales are fused by element-wise addition, patchified, and finally processed through transformer blocks with self-attention.

\paragraph{Post-Fusion Transformer}
We apply transformer blocks with both self-attention and cross-attention mechanisms. The fused YOLOv8 features are patchified (4×4 patches) and processed through 2 self-attention blocks followed by 2 cross-attention blocks. 
\begin{align}
    h'_{pr} &= \text{SelfAttn}^{(2)}(f_{pr}) \\
    h''_{pr} &= \text{CrossAttn}^{(2)}(Q=h'_{pr}, K=f_{wk}, V=f_{wk})
\end{align}
In the latter the queries are projections of the predictor embeddings, whereas the keys and values are projections of the DINO embedding. This procedure allows the detector representations to query foundation model semantic knowledge.

Self-attention refines YOLOv8 features by capturing long-range spatial dependencies across patches, while cross-attention allows the detector's representations to query DINO's semantic knowledge. This two-stage refinement creates semantically-grounded, spatially-coherent representations. We use 8 attention heads, 2 self-attention blocks, and 2 cross-attention blocks.

\paragraph{Dual Encoders}
Separate encoder heads map the refined features of the predictor and world-knowledge model to a shared 64-dimensional embedding space $\mathbb{R}^{64}$, which we denote as $e_{pr}$ and $e_{wk}$ respectively:
\begin{align}
    e_{pr} &= E_{PR}(h''_{pr}; \theta_{pr}) \in \mathbb{R}^{64} \\
    e_{wk} &= E_{WK}(f_{wk}; \theta_{wk}) \in \mathbb{R}^{64}
\end{align}
The YOLOv8 encoder processes the post-fusion transformer output through global pooling and projection. The DINO encoder is a deep 5-layer \ac{mlp}: $768 \to 1024 \to 768 \to 640 \to 512 \to 64$, with LayerNorm and \ac{gelu} activations.

\paragraph{Angular Failure Metric}
We measure this angle between the two embedings $e_{pr}$ and $e_{wk}$ of the two encoders via cosine similarity
\begin{equation}
    s_{\text{safety}}(\mathbf{x}) = \frac{e_{pr} \cdot e_{wk}}{\|e_{pr}\|_2 \|e_{wk}\|_2}.
\end{equation}
High similarity (small angle) indicates semantic similarity and predicted safety, while low similarity (large angle) signals misalignment and likely detector failure.

\subsection{Training Objective}
We train the model end-to-end using \ac{bce} loss. The cosine similarity score $s_{\text{safety}}(\mathbf{x}) \in [-1, 1]$ is mapped to safety probability $p_{\text{safe}}(\mathbf{x}) \in [0, 1]$ via $(1 - s)/2$, where -1 ($\pm180^\circ$ angle) maps to 1 (unsafe) and +1 ($0^\circ$ angle) maps to 0 (safe). \Ac{bce} loss is then applied with safety labels $y \in \{0, 1\}$ (0 = safe, all persons detected; 1 = unsafe, persons missed). With this encoding high cosine similarity (close embeddings) indicates safety, while low similarity (distant embeddings) indicates failure.

For model training and threshold tuning respectively we split COCO train 2017 (64,115 images) into training (90\%) and validation (10\%) sets \citep{lin2014microsoft}. COCO val2017 (2,693 images) serves as our held-out test set for final evaluation.

At test time, we compute the safety score $s_{\text{safety}}(\mathbf{x})$. The final prediction is based on a threshold, that is tuned on a disjoint validation set to 5\% \ac{fpr}. In our experiments of the full \ac{kgfp}, this threshold is set to $0.843$. That is, the probability for two random vectors in $64$ dimensions to have a cosine similarity above the threshold is negligible \citep{vershynin2018high}.
\section{Experimental Setup}

We evaluate \ac{kgfp} and established baselines on their ability to predict the correctness of YOLOv8's bounding boxes (\ac{iou} $> 0.5$). Both the object detector and visual foundation model backbones are frozen during all experiments.

\subsection{Datasets}

\paragraph{COCO 2017} We use only the person class from MS COCO~\citep{lin2014microsoft}: 64,115 training images and 2,693 validation images with 262,465 and 10,777 person annotations respectively. All other object classes are ignored, as we focus exclusively on detecting failures for safety-critical objects (persons). YOLOv8l is trained on COCO train split. Each image is labeled safe/unsafe based on whether YOLOv8l correctly detects all persons (\ac{iou} threshold 0.5, certainty threshold 0.5).

\paragraph{COCO-O Visual Domains} Following~\cite{hendrycks2021natural}, we evaluate on 6 \ac{ood} domains: \textit{cartoon} (artistic renderings), \textit{sketch} (line drawings), \textit{painting} (classical art), \textit{handmake} (crafts/toys), \textit{tattoo} (body art), and \textit{weather} (rain/snow/fog corruptions). These represent diverse visual anomalies ranging from stylistic variations (cartoon, sketch) to environmental corruptions (weather).
\begin{table*}[t]
\centering
\caption{Person Recall [\%] among \emph{accepted} images at 5\% FPR (selective prediction). Each method acts as a gate that rejects a fraction of images deemed unsafe; Person Recall is measured only on the images the gate accepts. \textit{YOLOv8 (base)} accepts all images (no gating). Best results per column in \textbf{bold}.}
\label{tab:main_results}
\small
\begin{tabular}{@{}lcccccccc@{}}
\toprule
\multirow{2}{*}{Method} & COCO & \multicolumn{6}{c}{COCO-O (Distribution Shift)} & COCO-O \\
\cmidrule(lr){3-8}
& Val & Cartoon & Sketch & Painting & Handmake & Tattoo & Weather & Avg \\
\midrule
\textbf{KGFP (Ours)} & \textbf{84.5} & \textbf{19.3} & 29.5 & \textbf{36.2} & 37.0 & \textbf{11.9} & \textbf{71.4} & \textbf{34.2} \\
\midrule
GRAM & 65.4 & 13.2 & 26.0 & 31.0 & 31.5 & 9.5 & 67.8 & 29.8 \\
KNN & 65.1 & 14.3 & 32.7 & 32.8 & \textbf{38.5} & 10.1 & 68.3 & 32.8 \\
ViM & 65.5 & 14.9 & \textbf{34.1} & 32.3 & 34.5 & 11.2 & 67.3 & 32.4 \\
\midrule
DINO-MLP & 69.1 & 13.6 & 26.3 & 33.4 & 33.1 & 8.2 & 65.2 & 30.0 \\
DINO-ViM & 65.2  & 15.4 & 34.0 & 32.1 & 34.8 & 7.5 & 66.6 & 31.7 \\
\midrule
\textit{YOLOv8 (base)} & 64.3 & 13.2 & 24.9 & 29.7 & 31.5 & 9.5 & 66.1 & 29.1 \\
\bottomrule
\end{tabular}
\end{table*}

\subsection{Implementation Details}

\paragraph{Model Configuration}
\Ac{kgfp} uses frozen YOLOv8l~\citep{jocher2023yolov8} and DINO (\ac{vit}-B)~\citep{caron2021dino} as feature extractors. YOLOv8l internal features are extracted from \ac{fpn} levels $\{P3/8, P4/16, P5/32\}$ at 640×640 resolution. DINO [CLS] tokens (768D) are extracted from 518×518 crops. The fusion architecture uses 8 attention heads with 2 self-attention blocks followed by 2 cross-attention blocks in the post-fusion transformer. Dual encoders project refined YOLOv8 and DINO features to a shared 64-dimensional embedding space (ablations test 128D, 256D, 512D). For efficiency, DINO embeddings are pre-computed to \ac{hdf} cache, while YOLOv8 features are computed on-the-fly during training.

\Ac{kgfp} is trained end-to-end using \ac{lars} optimizer~\citep{you2017large} with learning rate 0.00095, momentum 0.9, weight decay 0.0009, and \ac{lars} eta 0.001.
Training runs for 60 epochs with cosine annealing (T\_max=60, eta\_min=5e-7) and gradient clipping (max norm 1.0) for stability. We use batch size 6.

\subsection{Evaluation Metrics}
\textbf{Person Recall @ 5\% \ac{fpr}}: \textit{Primary safety metric} - percentage of ground-truth persons detected when accepting images where the safety score exceeds a threshold calibrated to yield 5\% \ac{fpr} on the \ac{id} validation split. Crucially, this single threshold is then applied unchanged to all COCO-O domains, simulating realistic deployment where no target-domain labels are available for recalibration. This metric directly quantifies safety: higher person recall means fewer missed persons in accepted images. Remark that if we deploy a random selective-prediction gate we expect the Person Recall @ 5\% \ac{fpr} to match the person recall of the original object detector, which serves as a random baseline for the metric. 

\textbf{\Ac{tpr} @ 5\% \ac{fpr}}: True positive rate (correct safety predictions on image level) at 5\% \ac{fpr}. Unlike Person Recall, this counts images correctly classified as safe/unsafe, not individual persons detected.

\textbf{\Ac{auroc}}: Area under \ac{roc} curve for binary safety classification (safe vs.\ unsafe images). \textbf{Rec \ac{auc}} and \textbf{Prec \ac{auc}}: Area under curve between the \ac{kgfp} module's \ac{fpr} and YOLO's Person Recall and Person Precision respectively on images predicted to be safe.

\subsection{Baselines}

In order to demonstrate the effectiveness of visual foundation models in object detector failure prediction, we compare our \ac{kgfp} against representative \ac{ood} detection methods adapted for object detection in safety monitoring. Since these methods were originally designed for image classification, we adapt them to work with YOLOv8l internal features extracted at multiple scales (P3, P4, P5). We train the \ac{ood} baseline methods GRAM, \ac{knn} and \ac{vim} using only safe images from the \ac{id} training set, treating unsafe images (where persons are missed) as anomalies in the classical \ac{ood} detection framing.

In order to support our proposed angular failure metric, we also compare our \ac{kgfp} to an \acp{mlp} baseline. Therefore, we train an ensemble of \acp{mlp} on the DINO embeddings. In our ablations we also train an \acp{mlp} head on DINO and YOLO embeddings as a baseline to our \ac{kgfp} (see Table~\ref{tab:all_ablations}).

\section{Results}

\subsection{Main Results}

\begin{table}[t]
\centering
\small
\caption{Comparison with OOD detection baselines. \textbf{Safety AUROC} measures binary classification (safe vs unsafe). \textbf{Person Rec. AUC} measures YOLOv8 Recall averaged over the FPR of KGFP. \textbf{Person Prec. AUC} measures YOLOv8 precision averaged over the FPR of \ac{kgfp}.}
\begin{tabular}{@{}l|ccc|ccc@{}}
\toprule
& \multicolumn{3}{c|}{\textbf{COCO Val}} & \multicolumn{3}{c}{\textbf{COCO-O Avg}} \\
\cmidrule(lr){2-4} \cmidrule(lr){5-7}
Method & Safety & Rec. & Prec. & Safety & Rec. & Prec. \\
& AUROC & AUC & AUC & AUROC & AUC & AUC \\
\midrule
\textbf{KGFP} & \textbf{92.9} & \textbf{90.4} & \textbf{98.2} & \textbf{80.3} & \textbf{55.8} & \textbf{94.7} \\
GRAM & 52.5 & 65.6 & 94.1 & 52.4 & 31.1 & 93.4 \\
VIM & 53.3 & 67.5 & 89.4 & 66.2 & 44.4 & 90.5 \\
KNN & 54.3 & 66.6 & 90.0 & 68.9 & 47.5 & 92.7 \\
\bottomrule
\end{tabular}
\label{tab:auroc_metrics}
\end{table}

\begin{figure}[t]
\centering
\includegraphics[width=\columnwidth]{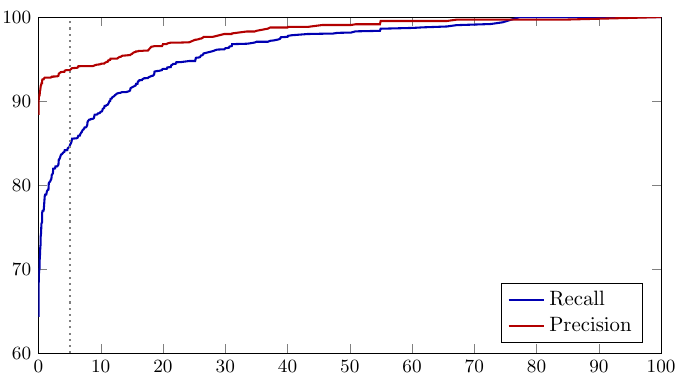}
\caption{KGFP performance on COCO Val. We plot person recall (blue) and precision (red) on the y-axis [in \%] versus KGFP FPR on the x-axis [in \%].}
\label{fig:yolo_curves}
\end{figure}

 Table~\ref{tab:main_results} presents our main results. \Ac{kgfp} acts as a selective-prediction gate: it rejects images deemed likely to contain missed persons and does not modify YOLOv8's detections themselves. Among the \emph{accepted} images, 84.5\% of ground-truth persons are correctly detected, compared to 64.3\% when accepting all images without gating. On the COCO-O weather domain, gating raises person recall among accepted images from $66.1\%$ to $71.4\%$ at $5\%$ \ac{fpr}.
This corresponds to a relative reduction of person oversights (YOLOv8 false negatives) of $56.6\%$ on COCO Val and $15.6\%$ on the COCO-O weather split at $5\%$ \ac{fpr} compared to $0\%$ \ac{fpr}.

\begin{table*}[t]
\centering
\caption{Ablation studies: (left) architecture components, (center) foundation model comparison, (right) embedding dimensions and optimizer. All values are Person Recall @ 5\% \acs{fpr}. \acs{id}: COCO Val; \acs{ood}: average over six COCO-O domains. Architecture abbreviations: pre-fn = pre-fusion, post-fn = post-fusion, attn = attention.}
\label{tab:all_ablations}
\small

\begin{minipage}[t]{0.33\linewidth}
\centering
\label{tab:ablation_architecture}
\begin{tabular}{@{}lcc@{}}
\toprule
\makecell[l]{Attention Ablation\\Head Ablation} & ID & OOD \\
\midrule
\textbf{Full KGFP} & \textbf{84.5} & \textbf{34.2} \\
\midrule
No pre-fn attn & 83.4 & 31.0 \\
No post-fn cross-attn & 83.4 & 32.5 \\
No post-fn self-attn & 84.2 & 34.1 \\
No attn & 82.3 & 31.1 \\
\midrule
MLP head & 83.7 & 34.2 \\
\bottomrule
\end{tabular}
\end{minipage}
\hfill
\begin{minipage}[t]{0.33\linewidth}
\centering
\label{tab:foundation_models}
\begin{tabular}{@{}lccc@{}}
\toprule
\makecell[l]{World-Knowledge\\Model (Parameter)} & ID & OOD \\
\midrule
\textbf{DINO (86M)} & \textbf{85.1} & \textbf{34.2} \\
\midrule
SigLIP (87M) & 84.8 & 35.2 \\
DINOv2 (304M) & 84.4 & 33.9 \\
CLIP (427M) & 84.5 & 32.6 \\
\bottomrule
\end{tabular}
\end{minipage}
\hfill
\begin{minipage}[t]{0.33\linewidth}
\centering
\label{tab:embedding_dims}
\begin{tabular}{@{}lcc@{}}
\toprule
\makecell[l]{Embedding\\ Dimension,\\Optimizer} & ID & OOD \\\midrule
\textbf{64D, Lars} & \textbf{84.5} & \textbf{34.2} \\
\midrule
64D, Adam & 84.3 & 32.8 \\
128D, LARS & 83.1 & 33.1 \\
256D, LARS & 83.5 & 31.1 \\
256D, Adam & 65.1 & 29.3 \\
512D, LARS & 67.1 & 29.1 \\
\bottomrule
\end{tabular}
\end{minipage}
\label{tab:ablations}
\end{table*}

We adapt \ac{ood} detection methods for failure prediction by training them on safe images only, treating unsafe images as \ac{ood}. On \ac{id} data, GRAM, \ac{knn}, and \ac{vim} achieve 65.1--65.5\% recall, only slightly above the YOLOv8 baseline (64.3\%). DINO-based variants (DINO-\ac{mlp}: 69.1\%, DINO-\ac{vim}: 65.2\%) achieve higher recall but remain below \ac{kgfp}. \Ac{kgfp} outperforms the best baseline by +15.4 percentage points, indicating that the dual-encoder architecture with explicit failure supervision is more effective than treating failures as \ac{ood} samples.

On COCO-O, \ac{kgfp} achieves the \textbf{highest average person recall (34.2\%)} across all six COCO-O domains. Note that \ac{ood} baselines were trained to treat unsafe images as \ac{ood} samples (see Section~4.4), so they face a more challenging task on COCO-O: distinguishing unsafe from safe images under genuinely novel visual conditions. \Ac{kgfp} performs notably better on \textit{weather} corruptions (71.4\%), maintaining near-\ac{id} performance. On \textit{painting} (36.2\%) and \textit{handmake} (37.0\%), \ac{kgfp} leads or matches the best baselines. These results indicate that the learned alignment between YOLOv8 and DINO representations generalizes to novel visual domains.

Table~\ref{tab:auroc_metrics} provides complementary area-under-curve metrics. \Ac{kgfp} achieves 92.9\% Safety \ac{auroc} for binary safe/unsafe classification and 90.4\% Person Rec.\ \ac{auc} on COCO Val, compared to 54.3\% Safety \ac{auroc} for the best baseline (\ac{knn}). On COCO-O, \ac{kgfp} maintains 80.3\% Safety \ac{auroc} and 55.8\% Person Rec.\ \ac{auc} on average across all novel visual domains.

\subsection{Ablation Studies}

Table~\ref{tab:all_ablations} (left) analyzes architectural components. Removing all attention modules reduces \ac{id} recall by 2.3\% (84.5\% $\rightarrow$ 82.3\%). Pre-fusion attention and post-fusion cross-attention each contribute $\sim$1\% to \ac{id} performance, while post-fusion self-attention has negligible impact. Replacing cosine similarity with an \ac{mlp} similarity head degrades Rec@5\%\ac{fpr} by 0.8\%. Mean cosine similarity remains below 0.9 throughout training, indicating no embedding collapse; deeper DINO embedding \acp{mlp} (7+ layers) collapsed in preliminary experiments.

Table~\ref{tab:all_ablations} (right) reports results across embedding sizes $d \in \{64,128,256,512\}$ and optimizers. The 64D embedding performs best on both COCO Val (84.5\%) and COCO-O (34.2\%) with only 2.6M parameters. Larger embeddings show diminishing returns, with clear overparameterization at 512D. \Ac{lars} slightly outperforms Adam~\citep{kingma2015adam} at 64D, while Adam performs poorly at 256D.
Table~\ref{tab:all_ablations} (center) compares world-knowledge encoders. DINO \ac{vit}-B/16 (86M parameters) achieves the highest \ac{id} recall (85.1\%) with the fewest parameters, outperforming CLIP \ac{vit}-L/14 (84.5\%, 427M parameters), DINOv2 \ac{vit}-L/14 and SigLIP \ac{vit}-B/16. We attribute this to DINO's self-distillation objective producing spatially coherent attention maps that capture fine-grained visual cues as partial occlusions, atypical poses, unusual lighting ~\citep{caron2021dino}, whereas CLIP's language-aligned features are shaped by caption-level semantics~\citep{radford2021clip} and thus less sensitive to these sub-textual failure patterns.

\subsection{Discussion}

\textbf{Supervised vs.\ unsupervised.} \Ac{kgfp} substantially outperforms all baselines on \ac{id} data. However, \ac{kgfp} is supervised (trained with safe/unsafe labels) while the \ac{ood} baselines are unsupervised (fitted on safe images only). A fairer comparison is with the supervised DINO-\ac{mlp} and the \ac{mlp}-head ablation (Table~\ref{tab:all_ablations}), which use the same labels. \Ac{kgfp} outperforms both, indicating that the dual-encoder cosine-similarity formulation provides a stronger inductive bias for failure prediction than direct classification, despite having fewer trainable parameters.

\textbf{Role of DINO features.} Methods based solely on DINO features (DINO-\ac{mlp}, DINO-\ac{vim}) perform worse than \ac{kgfp}, confirming that DINO embeddings alone cannot reliably predict YOLOv8 failures---the fusion of both feature sources is essential. DINO provides complementary semantic context (scene-level patterns, occlusion cues) that is correlated with detector failure but not directly accessible from YOLOv8's task-specific features.

\textbf{OOD generalization.} The unsupervised baselines show near-baseline performance on \ac{id} data, which is expected: failures on \ac{id} images do not necessarily correspond to distributional novelty. On COCO-O, unsupervised baselines achieve competitive or superior performance on specific domains---\ac{knn} on handmake (38.5\% vs.\ 37.0\%) and \ac{vim} on sketch (34.1\% vs.\ 29.5\%)---because genuine distribution shift correlates with detector failure in these stylistically extreme domains. However, \ac{kgfp} achieves the highest average recall across all six domains, demonstrating more consistent generalization.

\textbf{Embedding dimensionality.} The degradation at 256D and 512D (Table~\ref{tab:all_ablations}, right) is attributable to overparameterization: larger embedding spaces require more data to learn meaningful angular structure, and cosine similarity becomes less discriminative at high dimensionality because random vectors concentrate around orthogonality~\citep{vershynin2018high}. The 64D space provides sufficient capacity while maintaining a well-structured angular decision boundary.

\section{Limitations and Future Work}

\Ac{kgfp} requires frozen foundation model embeddings. Changes to the foundation model or encounters with visual domains far beyond its training data could degrade performance. Future work should investigate continual adaptation of world-knowledge encoders.
The all-or-nothing safety label (all persons detected vs.\ any missed) may be too coarse for some applications. Extensions could predict expected miss counts or provide spatial failure localization.
We focus on person detection for clear safety implications. Generalization to multi-class scenarios requires rethinking the safety formulation—which objects are safety-critical?
\Ac{kgfp} requires computing both YOLOv8 and DINO forward passes, which introduces significant overhead.
Latency is critical in real-time systems. Future work should investigate the latency of our architecture, how it could be reduced, and what the trade-offs would be.
The Latency is mainly caused by the backbones and the classification head introduces negligible latency.
Frame-skipping strategies (running safety checks every $N$ frames) can amortize costs for video streams, or deployment may benefit from model compression, distillation, or efficient foundation model variants.
We evaluate on naturally occurring visual anomalies. Adversarial perturbations designed to fool both YOLOv8 and DINO could evade safety monitoring. Adversarial training or certified defenses warrant investigation.

\section{Conclusion}

We introduced \ac{kgfp}, a runtime monitoring framework that detects unsafe images where a specific object detector misses safety-critical objects by measuring semantic alignment between learned encodings of object detector's internal activations and DINO embeddings. The architecture fuses multi-scale YOLOv8 features via pre-fusion cross-scale attention, then applies post-fusion cross-attention with DINO for semantic alignment measurement. \Ac{kgfp} provides actionable failure signals for safety-critical deployment. Focusing on safety-critical object classes (persons for pedestrian detection), \ac{kgfp} achieves 85\% person recall at 5\% \ac{fpr} on both \ac{id} data and novel visual domains, substantially outperforming traditional \ac{ood} detection methods and our angular failure metric outperforms an MLP baseline with fewer parameters.
Comprehensive ablations reveal that foundation model world-knowledge, cross-attention fusion, and angular metric learning are critical for robust failure prediction.
As object detectors proliferate in autonomous vehicles, surveillance, and healthcare, explicit failure monitoring becomes essential. \Ac{kgfp} demonstrates that foundation models trained on billions of images can serve as semantic "sanity checks" for task-specific detectors, identifying unsafe images before they cause harm.

\section*{Broader Impact and Disclosures}
\label{sec:impact_disclosure}

\paragraph{Ethics Statement.} KGFP is designed to enhance safety by detecting object detector failures. It functions as an additional safety layer, not a replacement for robust model development. Limitations include potential false negatives or positives; users must carefully consider recall-precision trade-offs and evaluate performance across relevant demographic groups to mitigate potential dataset biases.

\paragraph{LLM Usage.} We acknowledge the use of Claude 3.5 Sonnet and GitHub Copilot for writing refinement, coding assistance, and literature summarization. All novel algorithmic components, experimental designs, and scientific conclusions are the sole work of the human authors. The authors take full responsibility for all content, including any errors or inaccuracies that may have been introduced during LLM-assisted editing. 

{
    \small
    \bibliographystyle{ieeenat_fullname}
    \bibliography{cao_references}
}

\end{document}